\documentclass[letterpaper]{article} 
\usepackage{aaai2027}  
\usepackage[hyphens]{url}  
\usepackage{graphicx} 
\usepackage{natbib}  
\usepackage{caption} 
\usepackage{amsmath,amssymb}
\usepackage{booktabs}
\usepackage{multirow}
\usepackage{array}
\usepackage{adjustbox}
\usepackage{xcolor}
\usepackage{tcolorbox}
\usepackage{tikz}
\usetikzlibrary{arrows.meta,positioning,shapes.geometric,fit,backgrounds}
\definecolor{clinicalblue}{RGB}{33, 102, 172}
\definecolor{chunkgreen}{RGB}{27, 120, 55}
\definecolor{judgeorange}{RGB}{179, 88, 6}
\definecolor{panelgray}{RGB}{240, 240, 240}
\usepackage{colortbl}   
\IfFileExists{soul.sty}{\usepackage{soul}}{}
\makeatletter
\@ifundefined{sethlcolor}{%
  \newcommand{\hl@color}{hlA}%
  \newcommand{\sethlcolor}[1]{\renewcommand{\hl@color}{#1}}%
  \DeclareRobustCommand{\hl}[1]{{\setlength{\fboxsep}{1pt}\colorbox{\hl@color}{#1}}}%
}{}
\makeatother
\definecolor{hlA}{HTML}{CFE7E2}      
\definecolor{hlB}{HTML}{E5DBF1}      
\definecolor{hlC}{HTML}{F5E5C0}      
\definecolor{okInk}{HTML}{2E7D32}    
\definecolor{noInk}{HTML}{C0392B}    
\definecolor{hdrBand}{HTML}{EFEFEF}  
\definecolor{stubBand}{HTML}{F7F7F7} 
\DeclareRobustCommand{\hA}[1]{{\setlength{\fboxsep}{1pt}\sethlcolor{hlA}\hl{#1}}}
\DeclareRobustCommand{\hB}[1]{{\setlength{\fboxsep}{1pt}\sethlcolor{hlB}\hl{#1}}}
\DeclareRobustCommand{\hC}[1]{{\setlength{\fboxsep}{1pt}\sethlcolor{hlC}\hl{#1}}}
\DeclareRobustCommand{\vchip}[1]{\textcolor{#1}{\rule[0.15ex]{3.6pt}{3.6pt}}}
\DeclareRobustCommand{\vOK}{\vchip{okInk}\,\textcolor{okInk}{\textbf{correct}}}
\DeclareRobustCommand{\vNO}{\vchip{noInk}\,\textcolor{noInk}{\textbf{incorrect}}}
\DeclareRobustCommand{\fmode}[1]{\textit{\textcolor{black!62}{#1}}}
\newcommand{\rlab}[1]{\textsc{#1}}
\newcommand{\rowsep}{\arrayrulecolor{black!16}\midrule\arrayrulecolor{black}}
\newcolumntype{L}[1]{>{\raggedright\arraybackslash}p{#1}}
\newcolumntype{S}[1]{>{\columncolor{stubBand}\raggedright\arraybackslash}p{#1}}
\newcommand{\cmark}{\textcolor{chunkgreen}{\checkmark}}
\newcommand{\xmark}{\textcolor{judgeorange}{$\times$}}
\nocopyright
\title{MedUPS: Towards Diagnostic Assistance in Uncommon Medical Cases with Large Language Models}
\author{
    Ofir Ben Shoham\equalcontrib,
    Oriel Perets\equalcontrib,
    Nir Grinberg,
    Nadav Rappoport
}
\affiliations{
    Stein Faculty of Computer and Information Science\\
    Ben-Gurion University of the Negev\\
    \{benshoho, orielpe\}@post.bgu.ac.il, \{nigrn, nadavrap\}@bgu.ac.il
}
\begin{document}
\maketitle
\begin{abstract}
Uncommon and off-guideline cases are difficult for clinical decision support, because physicians must make a series of management decisions under diagnostic uncertainty and rarely see the full case at once. Most large language model (LLM) benchmarks for medicine score only the final diagnosis, yet much of clinical care turns on the next appropriate action: the next test to order, the imaging study to obtain, the specialist to involve, or the differential to pursue. We introduce MedUPSQA, a dataset of $21{,}874$ mid-stream clinical decision points built from $5{,}535$ real case reports, and MedUPS, an alignment framework that supervises models on these intermediate decisions as they unfold along a patient's trajectory. We segment free-text case presentations into chronologically ordered, accumulating clinical chunks and align models to predict the next step with reinforcement learning (GRPO), using an external LLM-as-a-Judge reward. This objective mirrors how clinicians actually meet patients, reasoning forward from accumulating evidence toward the next decision, rather than committing to a final label.
Across three backbones, mid-stream alignment raises next-step accuracy from $55.2$ to $66.7$ for Qwen3.6-27B, from $47.2$ to $57.8$ for Qwen3.5-9B, and from $37.8$ to $44.4$ for HuatuoGPT-3-8B, with 95\% CI. In several model scales we test the objective improves accuracy more than scale, with smaller models surpassing  larger, frontier models we evaluate. We further train supervised fine-tuning (SFT) baselines on the mid-stream task, SFT improves all backbones above base, indicating the target framwork carries signal independently of the optimizer. We release the dataset, code, and aligned checkpoints.
\end{abstract}
\section{Introduction}
Clinical large language models (CLLMs) now support a wide range of clinical tasks, including treatment prediction, medical coding, and diagnosis for both common and uncommon cases \cite{ben2024cpllm, lievin2024can, perets2025cupcase}. Their clinical value depends on more than accuracy; the underlying reasoning must be clinically coherent, and this matters most in complex or atypical presentations \cite{sonoda2025structured, wu2023large}.
The stakes are highest for uncommon and rare conditions. Diagnostic error is common and consequential: a landmark report estimates that most people will experience at least one diagnostic error in their lifetime, and that such errors are a leading source of preventable harm \cite{ball2015improving}. Among 6{,}507 patients with rare diseases, the average time from first medical contact to a confirmed diagnosis was 4.7 years, 73\% were misdiagnosed at least once, and 22\% consulted eight or more clinicians before receiving an answer \cite{faye2024time}. A diagnostic odyssey of this kind is produced by a long series of intermediate decisions made under uncertainty, such as which test to order, which specialist to involve, and which differential to pursue, each taken with only partial information in hand.
Clinicians reason the same way. They form and revise hypotheses as evidence accrues and choose the next action that best reduces uncertainty \cite{bowen2006educational}, and in real practice diagnoses and treatments unfold over time, requiring the integration of evolving evidence, prior interventions, and medical knowledge. Most medical LLM benchmarks instead score the terminal diagnosis from a fully specified vignette \cite{lievin2024can, perets2025cupcase}. This exam-style evaluation does not reflect the reality of clinical encounters, where clinicians must decide with incomplete information, and so provides limited evidence of a model's ability to support clinicians while a case is still unfolding.
For uncommon cases, the final diagnosis has several limitations as a training target. It arrives once per case, and it is recorded in free text whose phrasing varies from report to report. For rare presentations it is often uncertain even in the published record. We therefore supervise \emph{mid-stream} clinical decision-making: predicting the next appropriate clinical step along a patient's trajectory. A case that unfolds over $T$ chunks yields many such decision points, each with a concrete answer that the case itself later confirms, so the supervision is denser than a single terminal label and closer to what clinicians do at the bedside. We build MedUPSQA from real, segmented case presentations and align models to predict the next step using Group Relative Policy Optimization (GRPO) \cite{shao2024deepseekmath}, scored by an external LLM-as-a-Judge.
Our contributions are as follows:
\begin{enumerate}
    \item \textbf{A mid-stream alignment objective.} We formalize next-step prediction along patient timelines as a reinforcement learning objective and show that GRPO on this single objective improves next-step clinical reasoning on uncommon cases, across three unrelated backbones and with no diagnostic labels. We compare it against supervised fine-tuning on identical pairs, which isolates the contribution of the optimizer from that of the target.
    \item \textbf{MedUPSQA, an open dataset.} We release $21{,}874$ validated next-step decision points with reasoning traces, derived from segmented real-world case reports \footnote{ \url{https://huggingface.co/collections/oriel9p/medups}}, together with our aligned models and code\footnote{ \url{https://github.com/oriel9p/MedUPS}}. Throughout, MedUPSQA names the dataset and MedUPS names the alignment framework and the resulting model family (e.g., MedUPS-Qwen3.6-27B).
    \item \textbf{Alignment over scale.} Under our evaluation judge, mid-stream-aligned open models of modest size match or exceed substantially larger open and frontier models on next-step prediction, which suggests that targeted intermediate supervision can matter more than model size for uncommon-case reasoning. We quantify how much of this comparison is attributable to the identity of the judge, and are explicit about what that check does and does not settle.
\end{enumerate}
\begin{figure*}[t]
 \centering
 \includegraphics[width=\textwidth]{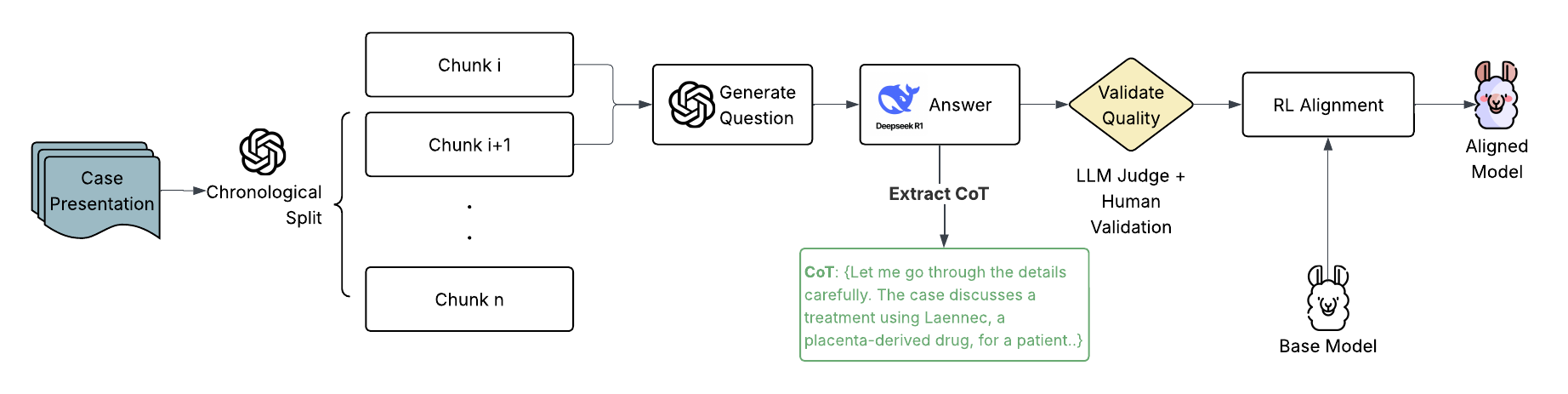}
 \caption{\textbf{MedUPSQA construction and mid-stream alignment.} Case reports are segmented into chronologically ordered chunks $c_1,\dots,c_T$. For each prefix, a question $q_i$ answerable from $c_{i+1}$ is generated with an answer $a_i$ and rationale $r_i$, and a judge keeps only instances clinically equivalent to $c_{i+1}$. The retained instances align policy models with GRPO. Every stage of construction is automated; no clinician adjudication is involved, which we treat as the principal limitation of the resource.}
 \label{fig:pipeline}
\end{figure*}
\section{Related Work}
A broad line of work applies reinforcement learning (RL) to clinical LLMs primarily as an alignment mechanism, aiming to match model behavior to clinician expectations and domain norms. Med-PaLM~2 optimizes a physician preference model and reports gains under criteria such as diagnostic prudence and reasoning clarity \cite{singhal2025medpalm2}, and, because expert supervision is expensive, \citet{yang2025rlaif_radiology_ejr} instead reward agreement with an AI evaluator to align radiology summaries with radiologist expectations. In these approaches, improvements in reasoning arise indirectly from preference or style alignment rather than from explicit supervision of clinical reasoning. A related group shapes rewards around clinical structure without an explicit reasoning objective: MMedPO for medical vision-language models \citep{zhu2024mmedpo, zhu2025mmedpo_icml}, QoQ~Med with hierarchy- and modality-aware reward scaling \citep{dai2025qoqmed}, and verifiable radiology reporting combining chain-of-thought with RL \citep{jing2025boxmedrl}.
More recent work uses RL to directly target reasoning behavior, typically by optimizing over reasoning traces or reward signals tied to verifiable outcomes. Med-RLVR demonstrates that RL over explicit reasoning traces can improve performance on medical multiple-choice benchmarks such as MedQA and MMLU-Pro Health \cite{zhang2025med}. Similarly, \citet{su2025crossing} extend verifiable-reward RL across domains, including medicine, by learning reward models that assign continuous scores in $[0,1]$ and rely on an auxiliary LLM as a judge for open-ended responses. While these methods indicate that RL can strengthen reasoning robustness, they are largely evaluated on common, clean, exam-style benchmarks with closed-form answers, and do not address longitudinal clinical decision-making or complex patient trajectories.
Sequential decision support predates LLMs. Reinforcement learning over electronic health records has been used to recommend the next clinical action directly from structured patient state, most prominently the AI Clinician, which learns sepsis treatment policies from ICU time series and evaluates them against clinician behavior \cite{komorowski2018aiclinician}. That literature operates on densely sampled structured variables within a narrow, guideline-covered condition, and learns from downstream outcomes. Our setting is complementary rather than competing: free-text presentations of uncommon conditions, where structured state is unavailable, outcome labels are absent, and the supervision signal is what the record says happened next.
A further line of work argues that static vignettes misrepresent clinical practice, and builds interactive settings in which information arrives over time. MediQ reformulates exam questions so that the model must recognize when its information is insufficient and ask follow-up questions before committing, and finds that strong LLMs seek information poorly even when they answer well under full information \citep{li2024mediq}. AgentClinic and CRAFT-MD instead simulate the encounter itself, pairing a model clinician with a simulated patient so that performance is measured over a dialogue rather than a one-shot answer, and both report substantial drops relative to vignette-style evaluation \citep{schmidgall2024agentclinic, johri2025craftmd}. DDXPlus supplies large-scale trajectories of evidence acquisition and differential diagnosis, but from a synthetic generative process rather than real encounters \citep{fansitchango2022ddxplus}. These resources establish that sequential, information-limited evaluation is both necessary and difficult. They are, however, evaluation environments: they measure off-the-shelf models and supply no training signal, and their cases are simulated or synthesized rather than drawn from real presentations. We take the same premise and turn it into supervision, deriving a dense training objective from the decision points that real case reports already contain.
The closest line of work improves diagnostic reasoning through explicit supervision of clinical rationales, and provides our main points of comparison. MedCaseReasoning assembles diagnostic cases paired with clinician-authored reasoning traces and reports that supervised fine-tuning (SFT) on these traces improves both diagnostic accuracy and reasoning recall \cite{wu2025medcasereasoning}. MedReason converts clinical question-answer pairs into step-by-step reasoning chains derived from a medical knowledge graph, and shows that SFT on the resulting chains raises medical reasoning across several 7--8B models \cite{wu2025medreason}. HuatuoGPT-o1 takes a two-stage route, first fine-tuning on verifier-guided reasoning trajectories and then applying PPO with a verifier-based reward \cite{chen2024huatuogpt}. All three depend on final-diagnosis labels or articulated rationales, and all three evaluate on exam-style or fully specified cases, which makes them costly to extend to uncommon presentations, rare diseases, and longitudinal settings where such labels are scarce or unreliable.
Our work departs from these methods in three ways. We supervise the intermediate next step along a patient's trajectory rather than the final diagnosis, so training needs no diagnostic labels. We optimize this objective with RL rather than SFT on curated rationales. And we build MedUPSQA on top of CUPCase \cite{perets2025cupcase}, a corpus of real, uncommon patient case reports, so both training and evaluation reflect the information-limited setting of an early clinical encounter rather than a clean benchmark. Table~\ref{tab:related} summarizes the comparison.
\begin{table*}[t]
\centering
\adjustbox{max width=\textwidth}{
\renewcommand{\arraystretch}{1.2}
\begin{tabular}{lcccccc}
\toprule
\textbf{Method} & \textbf{Prediction target} & \textbf{Training signal} & \textbf{Reasoning supervision} & \textbf{Setting / data} & \textbf{Dx labels} & \textbf{Trajectory-aware} \\
\midrule
MedReason \cite{wu2025medreason}          & MCQA answer       & SFT            & KG-derived chains  & Exam-style QA          & required & \xmark \\
MedCaseReasoning \cite{wu2025medcasereasoning} & Final diagnosis   & SFT            & Clinician traces   & Case reports           & required & \xmark \\
HuatuoGPT-o1 \cite{chen2024huatuogpt}     & Verifiable answer & SFT + RL (PPO) & Verifier-guided    & Exam-style QA          & required & \xmark \\
\midrule
\textbf{MedUPSQA (ours)}                   & \textbf{Next clinical step} & \textbf{RL (GRPO)} & \textbf{LLM-judge reward} & {Case reports} & \textbf{none} & \cmark \\
\bottomrule
\end{tabular}}
\caption{\textbf{Comparison with the three most closely related resources.} MedUPSQA differs in its prediction target, its reasoning supervision, and its independence from final-diagnosis labels, and in operating over accumulating evidence rather than a fully specified case.}
\label{tab:related}
\end{table*}
\section{Methodology}
Figure~\ref{fig:pipeline} gives an overview of the MedUPS pipeline. Where conventional alignment optimizes models directly for final-diagnosis prediction, our approach aligns them over the clinical trajectory by learning intermediate decision steps. Given a free-text case, we split it into temporally ordered clinical chunks and build supervision from prefixes of that trajectory, predicting the next clinically relevant step from the context available so far. This construction is denser by design: a case of $T$ chunks supplies $T-1$ supervised decisions where a final-diagnosis objective supplies one. We align models with a single objective, next-step prediction, using GRPO, and we perform no supervised fine-tuning and no training on the final-diagnosis label. We then evaluate the aligned models on a held-out set of mid-stream decision points.
\subsection{Problem Formulation}
Each case presentation is represented as an ordered sequence of clinical chunks $\{c_1, c_2, \dots, c_T\}$, where $c_i$ precedes $c_{i+1}$ in the clinical narrative. At decision point $i$, a policy model observes the accumulated context $\{c_1, \dots, c_i\}$ together with a question $q_i$ about the next clinically relevant step, and must produce the answer $a_i$ that is realized in the continuation $c_{i+1}$. This mirrors the information-limited setting of a real encounter, in which only part of the case has revealed itself and the clinician must decide what to do next.
\subsection{Dataset and Construction}
Our dataset extends CUPCase~\cite{perets2025cupcase} to $5{,}802$ medical case reports from the \emph{BMC Journal of Medical Case Reports}. Each case is a free-text clinical presentation; the accompanying free-text final diagnosis is not used anywhere in this work.
\paragraph{Chronological chunking.}
Each case presentation is split into a sequence of chronologically ordered, clinically coherent text chunks. Chunking uses few-shot prompting with GPT-4o, instructing the model to segment the case into minimal, self-contained units, each capturing a single dominant clinical concept (e.g., symptom evolution, physical examination finding, laboratory result, imaging report, medication change, or clinician assessment). This preserves temporal structure while avoiding conflation of multiple decision-relevant facts within one training instance. Chunking is fully automated and has not been verified by clinicians; the exact prompt and worked examples are released with the dataset.
\paragraph{Question and answer generation.}
Given the chunk sequence, we form instances from pairs $(\{c_1, \dots, c_i\}, c_{i+1})$, where the available patient context is all prior chunks up to step $i$ and supervision is derived from information introduced in $c_{i+1}$. We first generate a clinically meaningful question $q_i$ with an external LLM (DeepSeek-R1), grounded in the accumulated context and constructed so that it can be resolved from evidence in $c_{i+1}$. Conditioned on the same context and question, we then use DeepSeek-R1 in a few-shot setting to generate a chain-of-thought $r_i$ and a final answer $a_i$, giving instances $(\{c_1, \dots, c_i\}, q_i) \rightarrow (r_i, a_i)$.
\paragraph{Validation and splits.}
To ensure supervision quality, candidate questions and answers are screened by an external LLM judge, which sees the accumulated case context and determines whether the proposed answer is clinically equivalent to the reference realized in $c_{i+1}$, grading accuracy, completeness, contextual fit, and precision (full prompt in Appendix~\ref{app:prompts}, Figure~\ref{fig:prompt-validation}). We arrived at this prompt iteratively, manually inspecting $30$ judged instances after each revision and adjusting the criteria where the verdicts diverged from our reading of the reference. The review was carried out by the authors on the same instances that drove the revisions, so it is a development check on the prompt rather than an independent validation of the judge. Chunking and generation succeed on $5{,}789$ cases and yield $46{,}447$ candidate decision points; the filter retains $21{,}874$ ($47.1\%$) across $5{,}535$ cases, and these constitute MedUPSQA. The filter therefore discards slightly more than half of the generated supervision, trading dataset size for answers that are verifiably grounded in what the case reports next. To further ensure the quality, 4 authors randomly select 50 samples and validate their questions, and answers manually. We split at the case-report level \emph{before} chunking and question generation, assigning every decision point from a case to a single split, which gives $18{,}581$ train, $1{,}067$ validation, and $2{,}226$ test decision points with no case shared across splits. Full dataset statistics, including the distribution over question types and over position along the trajectory, are given in Appendix~\ref{app:dataset}, Table~\ref{tab:dataset-stats}.
\subsection{Reward}
We use an external LLM (DeepSeek-Chat) as a clinical judge to provide the reward signal during GRPO. Rather than a single binary correctness check, the judge grades each sampled answer with a four-criterion rubric, scoring each criterion on a $\{0,1,2\}$ scale: \textbf{Correct} (the answer matches the clinical action, finding, test, decision, or diagnosis realized in the reference continuation), \textbf{Specific} (the answer matches the reference's level of detail, neither too vague nor over-specified), \textbf{Grounded} (the answer is supported by the visible context and does not invent findings), and \textbf{Complete} (no essential component of a multi-part reference answer is missing). The criteria are combined with weights $(3, 1, 2, 1)$, chosen a priori to prioritize clinical correctness and evidence grounding and not tuned on validation data, into a raw score in $[0, 14]$ normalized to $[0, 1]$:
\begin{equation}
R(\hat{a}) = \tfrac{1}{14}\big(3\,s_{\text{corr}} + s_{\text{spec}} + 2\,s_{\text{ground}} + s_{\text{comp}}\big).
\end{equation}
We did not ablate these weights, and the sensitivity of the aligned models to them is unknown.
\subsection{Optimization}
We align models with Group Relative Policy Optimization (GRPO)~\cite{shao2024deepseekmath}, a critic-free policy-gradient method that normalizes each sampled completion's reward against the mean reward of its sampling group, avoiding a separately trained value network. For each prompt we sample a group of candidate completions, score each with the rubric judge, and update the policy to increase the likelihood of completions whose rewards exceed the group average. We perform parameter-efficient post-training with low-rank adapters (LoRA); only adapter parameters are updated, while a frozen reference model initialized from the same backbone provides KL regularization. Rollouts are generated with vLLM colocated and periodically re-synchronized with the updated adapter weights. The same optimization setup is used across all model families.
\begin{table*}[t]
\centering
\setlength{\tabcolsep}{4pt}
\renewcommand{\arraystretch}{1.12}
\scriptsize
\begin{tabular}{@{}S{0.095\linewidth} L{0.232\linewidth} L{0.232\linewidth} L{0.245\linewidth} L{0.105\linewidth}@{}}
\toprule
\rowcolor{hdrBand}
\textbf{Decision} & \textbf{Question} & \textbf{Prediction} & \textbf{Report} & \textbf{Verdict} \\
\midrule
\rlab{Diagnosis} \newline \textit{case 152}
 & What will be the most likely diagnosis given the patient's symptoms and ingestion of \hA{rhododendron} flowers?
 & \hB{Grayanotoxin} (rhododendron) poisoning.
 & Acute \hB{intoxication} by some species of \hA{rhododendron} was suspected.
 & \vOK \\
\rowsep
\rlab{Management} \newline \textit{case 3749}
 & What will be the next step in management given the persistent severe \hA{hyperkalemia} despite initial treatment?
 & Urgent \hB{hemodialysis} \textemdash{} potassium 8.9\,mmol/L, idioventricular rhythm, anuria, no response to shifting therapy.
 & The \hA{antikalemic} treatment was repeated without success, leading to an urgent consultation with a nephrologist for emergency \hB{dialysis}.
 & \vOK \\
\rowsep
\rlab{Management} \newline \textit{case 640}
 & What will be the next step in \hA{anticoagulation} management following recovery of the peripheral blood cell count?
 & Resume \hB{warfarin}; direct oral anticoagulants are contraindicated with mechanical valves.
 & \hA{Heparin} was changed to \hB{warfarin} and the patient was discharged from hospital.
 & \vOK \\
\rowsep
\rlab{Workup} \newline \textit{case 4950}
 & What will be the next diagnostic step for the painless mass on the lateral side of the left knee?
 & \hB{Magnetic resonance imaging} to characterize the lesion's size, depth and tissue composition in known Mazabraud syndrome.
 & Plain film radiographs did not show abnormalities; an \hC{ultrasound} examination showed a homogeneous, hypoechogenic soft-tissue mass with a cyst-like aspect.
 & \vNO \newline \fmode{ordering (C)} \\
\rowsep
\rlab{Pathology} \newline \textit{case 5529}
 & What will be the expected histopathological findings on biopsy of the rectal lesions?
 & Given long-standing \hA{Sj\"ogren's syndrome} and its known B-cell lymphoma risk, diffuse infiltration of small-to-medium lymphoid cells consistent with \hB{MALT lymphoma}.
 & \hC{Inflamed mucosa} with elongated tortuous glands attenuated toward the surface, and fibromuscular obliteration; the intervening mucosa was histologically normal.
 & \vNO \newline \fmode{clinical (E)} \\
\bottomrule
\end{tabular}
\caption{\textbf{Five decisions from the evaluation pool.} Shading links matching content between the prediction and the report. The two failures differ in kind: \textit{case 4950} disagrees only about the \emph{order} of the workup, the predicted MRI being performed later in the same case, while \textit{case 5529} is a genuine clinical error.}
\label{tab:qualitative-pool}
\end{table*}
\section{Experiments}
\paragraph{Models and adaptation.}
We apply mid-stream GRPO alignment to three open instruction-tuned backbones spanning $8$B to $27$B parameters: \textbf{Qwen3.6-27B}, \textbf{Qwen3.5-9B}, and the medical model \textbf{HuatuoGPT-3-8B}. We additionally report two unaligned open baselines under the same protocol, the medical reasoning model \textbf{MedReason-8B} and the substantially larger mixture-of-experts model \textbf{Qwen3.5-397B-A17B}. As non-aligned reference points, we evaluate strong frontier models, \textbf{GPT-5.6} (variants \textit{sol} and \textit{luna}), using identical prompts, answer extractors, and judge. All aligned models use \textbf{LoRA} adapters with base weights frozen. Exact model snapshots and access dates are listed in the released configuration files.
\paragraph{Supervised fine-tuning baseline.}
To separate the training \emph{target} from the \emph{optimizer}, we additionally fine-tune each of the three backbones on the identical mid-stream pairs with a standard supervised objective, using the generated Chain of Thought (CoT), and answer $(r_i, a_i)$ as the target sequence. These checkpoints are denoted with an \textsc{-sft} suffix. They see exactly the same training instances as the GRPO runs and use the same LoRA configuration, prompt template, answer extractor, and evaluation pool; the only difference is the learning signal. We did not tune the SFT recipe per backbone.
\paragraph{Prompting.}
All policy runs use a structured reasoning prompt that enforces a fixed output schema: the model reasons within a thinking span and emits its final answer inside a dedicated \texttt{<final\_answer>}\dots\texttt{</final\_answer>} span, from which the prediction is parsed. The system prompt and the next-step user template are reproduced in Appendix~\ref{app:prompts}, Figure~\ref{fig:prompt-policy}.
\paragraph{Reward and evaluation judges.}
The GRPO reward is the four-criterion rubric score defined above. At \emph{evaluation} time we use a stricter binary equivalence judge (DeepSeek-Chat) that scores a prediction correct only when its clinical content matches the reference; its prompt is given in Appendix~\ref{app:prompts}, Figure~\ref{fig:prompt-judge}.
\paragraph{GRPO alignment loop.}
Each optimization step follows a fixed pipeline: (i)~\textbf{Generate}, sample a group of $G{=}8$ completions per prompt with vLLM from the current policy; (ii)~\textbf{Extract \& judge}, parse the \texttt{<final\_answer>} span and score each completion with the rubric judge to obtain a reward in $[0,1]$; (iii)~\textbf{Advantage}, center each completion's reward against its group mean to form the GRPO advantage; (iv)~\textbf{Update}, apply the clipped GRPO policy-gradient update to the LoRA parameters with a KL penalty against the frozen reference. The optimizer, sampling, judge, and checkpoint-selection settings are identical across model families and are documented in full in Appendix~\ref{app:training}, with the full hyperparameter list in Table~\ref{tab:hparams}.
\paragraph{Evaluation protocol.}
All reported accuracies are LLM-judge equivalence rates on a fixed evaluation pool of $500$ decision points drawn from the $2{,}226$-item test split, stratified by the number of context chunks available at prediction time (1 through 8) so that positions along the trajectory are covered. Every model and every judge in this paper scores the same $500$ items, so all comparisons are made on identical instances. We report the bootstrap mean over $5$ resamples of that pool (seeds $1001$--$1005$) with a $95\%$ $t$-based interval. This interval quantifies sampling variation \emph{within} the evaluation pool only: each model was trained once, from a single seed, and evaluated at a single validation-selected checkpoint, so it bounds neither training nor decoding variance. We therefore describe results in 95\% CI.
\paragraph{Guarding against judge bias.}
The training reward and the primary evaluation metric are both produced by the same model family. An aligned policy could therefore raise its score by matching the idiosyncrasies of that judge while leaving its clinical decisions unchanged, and the comparison against frontier models is the most exposed to this concern, because those models were never optimized against our reward. We do not have clinician labels for these predictions, so we cannot establish which judge is closer to clinical assessment; what we can establish is how far a reported score moves when the judge changes. We therefore re-score the evaluation pool with two further automated judges, GPT-5.6 and Opus~5, under an identical prompt template, temperature, and answer extractor.
\section{Results}
\label{sec:results}
\begin{table*}[t]
\centering
{
\begin{tabular*}{\textwidth}{@{\extracolsep{\fill}}lcc@{}}
\toprule
\textbf{Model} & \textbf{0-shot} & \textbf{5-shot} \\
\midrule
\multicolumn{3}{l}{\textit{Open source models (base)}} \\
\midrule
Qwen3.5-397B-A17B                    & 53.16 {\scriptsize[50.23, 56.09]} & 47.00 {\scriptsize[45.20, 48.80]} \\
Qwen3.6-27B                          & 55.24 {\scriptsize[53.04, 57.44]} & 51.10 {\scriptsize[48.20, 54.00]} \\
Qwen3.5-9B                           & 47.24 {\scriptsize[44.21, 50.27]} & 42.40 {\scriptsize[40.80, 44.00]} \\
HuatuoGPT-3-8B                       & 37.80 {\scriptsize[35.61, 39.99]} & 39.80 {\scriptsize[38.20, 41.40]} \\
MedReason-8B                         & 35.92 {\scriptsize[33.91, 37.93]} & 32.10 {\scriptsize[30.00, 34.30]} \\
\midrule
\multicolumn{3}{l}{\textit{Frontier models (no alignment)}} \\
\midrule
GPT-5.6 \textit{sol}                 & 60.88 {\scriptsize[57.05, 64.71]} & 60.36 {\scriptsize[58.34, 62.38]} \\
GPT-5.6 \textit{luna}                & 59.76 {\scriptsize[57.05, 62.47]} & 59.08 {\scriptsize[57.18, 60.09]} \\
\midrule
\multicolumn{3}{l}{\textit{MedUPS models (ours): SFT vs.\ GRPO on identical pairs}} \\
\midrule
MedUPS-Qwen3.6-27B-SFT               & 59.60 {\scriptsize[58.40, 60.90]} & 58.40 {\scriptsize[55.80, 60.90]} \\
\textbf{MedUPS-Qwen3.6-27B} (GRPO)   & \textbf{66.68} {\scriptsize[64.09, 69.27]} & \textbf{59.90} {\scriptsize[56.20, 63.60]} \\
MedUPS-Qwen3.5-9B-SFT                & 55.60 {\scriptsize[54.90, 56.20]} & 53.20 {\scriptsize[51.40, 55.00]} \\
\textbf{MedUPS-Qwen3.5-9B} (GRPO)    & 57.80 {\scriptsize[55.56, 60.04]} & 52.20 {\scriptsize[49.40, 54.90]} \\
MedUPS-HuatuoGPT-3-8B-SFT            & 50.00 {\scriptsize[47.90, 52.20]} & 48.80 {\scriptsize[45.70, 52.00]} \\
\textbf{MedUPS-HuatuoGPT-3-8B} (GRPO)& 44.40 {\scriptsize[42.42, 46.38]} & 47.40 {\scriptsize[46.20, 48.50]} \\
\bottomrule
\end{tabular*}}
\caption{\textbf{Mid-stream next-step prediction accuracy (\%)} under the DeepSeek equivalence judge, on the fixed $500$-item evaluation pool. Bootstrap mean with a $95\%$ CI over $5$ resamples. Each \textit{MedUPS} checkpoint shares the backbone of the same name in the base block, and the \textsc{-sft} and GRPO variants of a backbone are trained on identical mid-stream pairs. Absolute values are judge-relative (see ``How much depends on the judge'') and are a lower bound on clinical competence (see ``Error analysis'').}
\label{tab:main_results}
\end{table*}
\paragraph{What a decision point looks like.}
Table~\ref{tab:qualitative-pool} shows five decisions drawn from the evaluation pool, chosen to span the question types and to include one ordering disagreement and one clinical error, with the question asked, the prediction, and the continuation the report recorded. The comparison is free text against free text, which is why grading requires a judge rather than string matching. Table~\ref{tab:qualitative} in Appendix~\ref{app:examples} follows a single uncommon case through three successive cuts, and its third cut previews the failure mode we return to below: the prediction is clinically reasonable but the reference records a different \emph{kind} of step.

\paragraph{Mid-stream alignment yields large gains.}
Table~\ref{tab:main_results} reports next-step accuracy. Mid-stream GRPO raises Qwen3.6-27B from $55.24$ to $66.68$ ($+11.4$ points), Qwen3.5-9B from $47.24$ to $57.80$ ($+10.6$), and HuatuoGPT-3-8B from $37.80$ to $44.40$ ($+6.6$). In every case the aligned results are higher than base in 95\% CI. The effect holds at two scales of a general-purpose backbone (Qwen) and on a domain-specialized medical backbone (HuatuoGPT), which is the strongest evidence we can offer that the gain reflects the objective rather than a quirk of one model family. Among the unaligned open models, the two trained on curated rationales for exam-style question answering, MedReason-8B ($35.92$) and HuatuoGPT-3-8B ($37.80$), score lowest, which fits our framing of the mid-stream setting as poorly covered by exam-style supervision, though these backbones differ from the Qwen models in other ways too.
\paragraph{RL beats supervised fine-tuning where the backbone is strong.}
Because the SFT and GRPO checkpoints of a backbone are trained on identical pairs, the comparison isolates the learning signal. SFT on MedUPSQA already lifts every backbone above its base, which confirms that the mid-stream target carries signal independently of how it is optimized. On the largest backbone GRPO is clearly the better of the two: MedUPS-Qwen3.6-27B reaches $66.68$ against $59.60$ for its SFT counterpart, and it is the only configuration in the table that overtakes the frontier systems. The advantage does not extend down the scale range. At $9$B the GRPO and SFT intervals overlap, and on HuatuoGPT-3-8B the SFT checkpoint is higher of the GRPO one; the same ordering holds under demonstrations, where GRPO leads only at $27$B. We read this as suggesting that group-relative RL pays off where the base policy is already strong enough for its sampled completions to differ informatively, and that on a weaker backbone imitating the reference answers is the safer signal. Since we did not tune the SFT recipe per backbone, part of this pattern may reflect recipe fit rather than a property of the objective.
\begin{figure}[t]
\centering
\includegraphics[width=0.9\columnwidth]{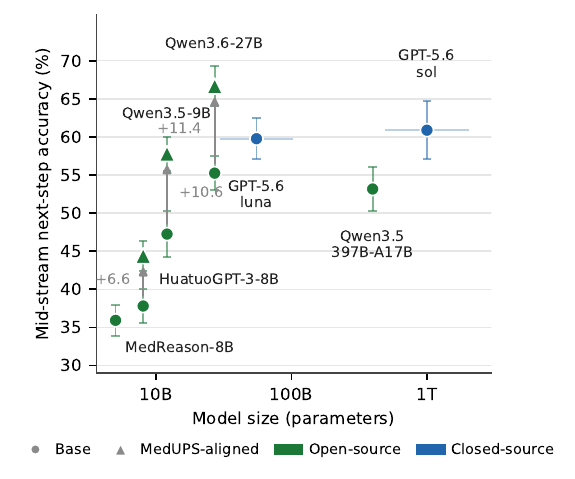}
\caption{\textbf{Mid-stream accuracy against model scale (0-shot).} Circles mark base models, triangles their MedUPS-aligned counterparts, and arrows the alignment gain. Open models (green) sit on the parameter axis. Frontier models (blue) appear in a separate panel sharing only the accuracy axis: their parameter counts are undisclosed, so no horizontal position should be read from them. Error bars are $95\%$ bootstrap CIs over the evaluation pool.}
\label{fig:scale}
\end{figure}
\begin{figure*}[t]
\centering
\includegraphics[width=\textwidth]{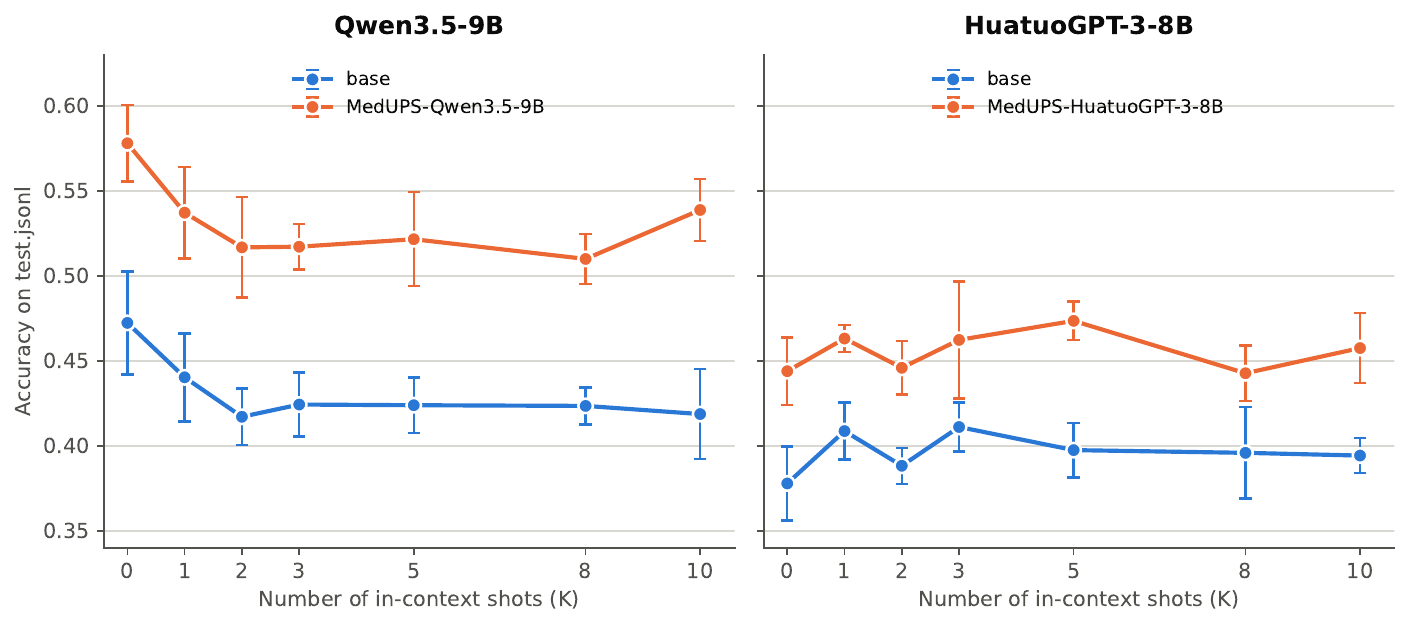}
\caption{\textbf{Few-shot ablation.} Mid-stream accuracy against the number of answer-only in-context demonstrations. Each aligned checkpoint stays above its own backbone at every shot count.}
\label{fig:fewshot-ablation}
\end{figure*}
\paragraph{Alignment can substitute for scale on this task.}
After mid-stream alignment the open Qwen3.6-27B ($66.68$) exceeds both GPT-5.6 variants (\textit{sol} $60.88$, \textit{luna} $59.76$), while the same backbone before alignment ($55.24$) trails them by roughly five points. The pattern repeats inside the open family: the aligned $9$B model reaches $57.80$, above the unaligned $27$B backbone at $55.24$ and within two points of GPT-5.6 \textit{luna}, so alignment recovers a roughly threefold parameter gap. Over the range of scales we test, the training objective separates the systems more than model size does; Figure~\ref{fig:scale} plots accuracy against parameter count, with frontier systems on a separate axis because their parameter counts are undisclosed. This comparison is also the one most exposed to judge identity, which we quantify next.
\paragraph{In-context demonstrations do not substitute for alignment.}
The $5$-shot column of Table~\ref{tab:main_results} tests whether the same behavior can be elicited by prompting instead of training. It cannot: five answer-only demonstrations lower accuracy for every open model and for both frontier variants, and the aligned checkpoints move in the same direction. The one exception is HuatuoGPT-3-8B, the weakest general instruction follower in the set, which gains $2.0$ points from having the output format demonstrated. The likely reason for the decline is that our demonstrations show the answer without the reasoning, which pulls models with a native thinking span toward short, under-reasoned outputs; the effect is largest where the $0$-shot model reasons most. This makes the $5$-shot column a floor on what prompting can achieve rather than a fair elicitation ceiling, since we did not test demonstrations that include chain-of-thought traces. The ordering is nonetheless unchanged: every trained checkpoint, SFT or GRPO, stays above its own base at $5$ shots. Figure~\ref{fig:fewshot-ablation} sweeps the shot count and shows the same picture across the range: demonstrations are drawn from a held-out shot pool and rendered answer-only, the demonstration items themselves are excluded from the evaluation pool, and each aligned checkpoint stays above its own backbone at every shot count.
\begin{table}[t]
\centering
\adjustbox{max width=\columnwidth}{
\begin{tabular}{lccc}
\toprule
\textbf{Judge pair} & \textbf{$n$} & \textbf{Agreement} & \textbf{Cohen's $\kappa$} \\
\midrule
DeepSeek / GPT-5.6-\textit{luna} & $500$ & $0.79$ & $0.56$ [$0.49$, $0.63$] \\
DeepSeek / GPT-5.6-\textit{sol}  & $500$ & $0.77$ & $0.53$ [$0.45$, $0.60$] \\
DeepSeek / Opus~5                & $500$ & $0.85$ & $0.65$ [$0.57$, $0.72$] \\
\bottomrule
\end{tabular}}
\caption{\textbf{Judge robustness.} The same $500$ predictions of MedUPS-Qwen3.6-27B scored independently by each judge under an identical prompt template and answer extractor. Agreement is the raw proportion of matching verdicts; $\kappa$ is Cohen's chance-corrected coefficient \cite{cohen1960coefficient} with a $95\%$ CI from its analytic standard error. Measured accuracy on this pool is $0.672$ under the DeepSeek judge and $0.580$ under GPT-5.6.}
\label{tab:judge_agreement}
\end{table}
\paragraph{How much depends on the judge.}
Neither judge is validated against clinician labels, so this check measures whether a reported score depends on which model produced the verdict, not whether either judge is right. Table~\ref{tab:judge_agreement} re-scores the evaluation pool with GPT-5.6 and Opus~5. DeepSeek and GPT-5.6-\textit{luna} agree on $79.2\%$ of instances ($\kappa=0.56$), moderate agreement on the bands of \citet{landis1977measurement}, and the disagreement is asymmetric: GPT-5.6 rejects $75$ predictions the DeepSeek judge accepted against $29$ in the reverse direction, so accuracy on this pool falls from $0.672$ to $0.580$. Opus~5 sits closer to DeepSeek ($0.85$ agreement, $\kappa=0.65$), so the DeepSeek--GPT-5.6 gap is the widest of the three pairings rather than a typical one. Roughly one prediction in five sits where two capable models disagree about clinical equivalence, which is a property of the task rather than of either judge.
We re-scored our strongest checkpoint rather than every system in Table~\ref{tab:main_results}, so this check bounds the \emph{magnitude} of the judge effect but does not verify that the model ordering survives a judge swap. The shift we measure, $9.2$ points, is larger than the $5.8$-point margin by which MedUPS-Qwen3.6-27B leads GPT-5.6 \textit{sol}; we therefore do not claim that the open-versus-frontier ordering is judge-independent, and re-scoring the full table under a second judge family is the follow-up we regard as most necessary. The aligned-versus-base comparisons are exposed in the opposite direction, since only the aligned model was optimized against a DeepSeek-family reward. What argues against those gains being pure judge matching is their consistency across three unrelated backbones and the error analysis below, in which the surviving failures are dominated by properties of the reference rather than of the judge.
\paragraph{Error analysis.}
A graded failure on this task is not automatically a clinical failure. The reference is whatever one published report happened to record next, so a prediction can be marked incorrect for reasons that have nothing to do with the model's reasoning. To separate the two, we inspected every failure of MedUPS-Qwen3.6-27B on the evaluation pool and assigned each to one of six categories: (A)~the reference is an under-determined particular of a single patient that no reasoner could recover from the visible context; (B)~the reference does not address the question asked; (C)~the prediction names a defensible step at the wrong point in the workup or the wrong granularity; (D)~question and continuation concern different \emph{kinds} of step; (E)~the prediction is clinically wrong on the information available; (F)~the judge erred. Assignment was performed by an LLM rater instructed to resolve doubt \emph{against} calling a failure clinical, so category~E is a lower bound on genuine reasoning error, and it was not verified by clinicians.
Crossing this attribution with a stratification by question stem shows the two dominant strata failing for opposite reasons. \textsc{Action} questions, which ask what to do next, are the hardest stratum ($58.7\%$ against $68.6\%$ for \textsc{Finding}), yet not one of their failures is a clinical error and roughly three quarters are disagreements about which of several defensible next steps the report happened to take first. \textsc{Finding} questions instead fail because the reference is an arbitrary particular of one patient (A) or does not answer the question (B), which together account for about four fifths of that stratum's failures. Aggregated over the pool, category~E covers $2.0\%$ of all decisions.
Two consequences follow for how Table~\ref{tab:main_results} should be read. The headline accuracies understate clinical competence by an amount that varies systematically with question type, so differences between models are informative while the absolute level is not a measure of clinical safety. And the largest single source of graded failure on \textsc{Action} questions, the class with the most direct bedside analogue, is under-specification of the reference rather than model error, which is a concrete argument for the multi-reference standard we flag below. We ran this attribution only for our strongest aligned model, so we cannot rule out that the failure composition differs for the frontier baselines, which would change how the cross-system margins should be interpreted.
\section{Discussion}
\label{sec:discussion}
\paragraph{Chunks as a patient trajectory.}
Splitting a case into chronologically ordered chunks turns a static vignette into a sequence of decision points where information grows step by step, which is closer to a real encounter than a fully specified case. Asking the model for the next step at each point rewards what clinicians value: reasoning forward from what is known so far, choosing the most informative next action, and staying calibrated when the picture is incomplete. It also gives denser supervision than a single terminal label, since a case with $T$ chunks yields many next-step decisions, each with a concrete answer that can be checked against what the case reports next.
\paragraph{Alignment over scale, and why it matters for deployment.}
A mid-stream-aligned $27$B open model surpasses the frontier systems we tested under our judge, and an $8$B medical model improves sharply under the same recipe. Open models of this size can run on-premises inside hospital infrastructure, which eases the privacy, cost, and auditability constraints that make closed frontier APIs hard to use on patient data. Subject to the judge caveat above, our results point to the alignment objective, rather than ever-larger closed models, as a workable direction for clinically useful systems. We did not measure inference latency or serving cost, so the operational side of that argument remains untested.
\paragraph{Why uncommon cases need dedicated models.}
Rare and off-guideline presentations are where general systems tend to fail and where patients are most often harmed \cite{ball2015improving, faye2024time}. These cases are underrepresented in training corpora and often fall outside clinical guidelines, so a model tuned for common, guideline-covered conditions cannot be assumed to carry over. Building and evaluating on uncommon-case trajectories, as MedUPSQA does, targets the population with the greatest unmet need.
\paragraph{Ethical considerations.}
MedUPS is intended as clinical decision support under clinician oversight, and over-reliance on its next-step suggestions is a real risk that deployment must guard against. Supervision and evaluation both rely on an LLM-as-a-Judge, which could carry errors and biases; we reduce this with a strict rubric, score caching for consistency, a separate binary evaluation judge, and two additional judge families, we recommend expert audit before clinical use. The underlying case reports come from the published literature and carry its publication, geographic, and demographic biases; we report the observable composition of the corpus in Appendix~\ref{app:dataset}, but the reporting in case reports is incomplete enough that we cannot characterize which patient populations the resulting models are and are not validated for. This is a material limitation for any clinical use. We release the data, code, and checkpoints so that others can, replicate, and strengthen these safeguards.
\paragraph{Limitations and future work.}
Our reward and our primary metric are produced by language models, and neither is validated against clinician labels; the $30$-instance review described above shaped the judge prompt during development but was neither held out nor performed by independent clinicians, so it does not stand in for such validation, and expert annotation of a held-out sample is the natural next step and the one we regard as most needed. Every stage of dataset construction, filtering, reward assignment, evaluation, and error attribution runs through LLMs, so the retained supervision is by definition what an automated judge considered equivalent, and cumulative bias cannot be excluded. We report rank stability under an alternate judge for our strongest checkpoint, as discussed above. We compare GRPO against supervised fine-tuning on identical mid-stream pairs, but not against applying the same GRPO recipe to final-diagnosis labels, so we establish what the mid-stream target buys relative to how it is optimized rather than relative to the conventional training target. The mixed GRPO-versus-SFT outcome across backbones is itself unexplained, and because the SFT recipe was not tuned per backbone we cannot separate a property of the objective from a property of the recipe. Further ablations would strengthen the paper: rubric weights, group size $G$, the LoRA rank, and the training seed. Our cases come from a single case-report corpus, which bounds their diversity and leaves generalization to other journals, however, due to the open, worldwide access to the journal, and the time frame of reports, we assume diversity is present. Because that corpus is public, contamination of the pretraining data of the models we compare cannot be excluded, however, the small representation of the specific case reports in the incredibly large pretraining data should mitigate this limitation slightly. Finally, a reference next step drawn from what published reports happened to do is one defensible action rather than the only one, the retrospective nature of the report should strenghten this action as a solid choice in the patient trajectory. Useful next steps include a reference standard admitting multiple acceptable next steps, calibration and abstention when the next step is genuinely uncertain, a breakdown of accuracy as a function of position along the trajectory, and multimodal trajectories that add imaging and laboratory data.
\section{Conclusion}
We introduced MedUPSQA, a dataset of $21{,}874$ mid-stream clinical decision points, and MedUPS, an alignment framework that supervises language models on the intermediate clinical decisions unfolding along a patient's trajectory instead of on final diagnostic labels. Aligning models on next-step prediction with GRPO produced large gains on mid-stream reasoning across three backbones spanning $8$B to $27$B parameters, with separated bootstrap intervals in every case. Supervised fine-tuning on the identical pairs also lifts every backbone, so the target carries signal independently of the optimizer, while GRPO is the stronger of the two at the largest scale we test. Under our evaluation judge, a mid-stream-aligned open model also surpassed substantially larger frontier systems, and a $9$B aligned model surpassed its own unaligned $27$B counterpart, which suggests that for uncommon cases the alignment objective can carry more weight than scale over this range. Because both the reward and the metric are automated, we are explicit about what the evaluation does not settle, and we release our dataset, code, and checkpoints to support the clinician-grounded validation that this line of work now requires.
\bibliography{aaai2027}
\appendix
\setcounter{figure}{0}\renewcommand{\thefigure}{A\arabic{figure}}
\setcounter{table}{0}\renewcommand{\thetable}{A\arabic{table}}
\section{Dataset Statistics}
\label{app:dataset}
Table~\ref{tab:dataset-stats} reports the composition of MedUPSQA, including the distribution over question types and over position along the case trajectory.
\begin{table}[t]
\centering
\adjustbox{max width=\columnwidth}{
\begin{tabular}{lr}
\toprule
\textbf{Statistic} & \textbf{Value} \\
\midrule
\multicolumn{2}{l}{\textit{Construction}} \\
Case reports (CUPCase extension)            & $5{,}802$ \\
Cases successfully chunked and generated    & $5{,}789$ \\
Candidate decision points                   & $46{,}447$ \\
\midrule
\multicolumn{2}{l}{\textit{After equivalence validation}} \\
Retained decision points                    & $21{,}874$ ($47.1\%$) \\
Cases represented                           & $5{,}535$ \\
\midrule
\multicolumn{2}{l}{\textit{Splits (disjoint at case level)}} \\
Train                                       & $18{,}581$ \\
Validation                                  & $1{,}067$ \\
Test                                        & $2{,}226$ \\
\midrule
\multicolumn{2}{l}{\textit{Per-case structure}} \\
Chunks per case after processing (mean)             & 4 \\
\bottomrule
\end{tabular}}
\caption{\textbf{MedUPSQA dataset statistics.} Splits are disjoint at the case-report level: every decision point derived from a case is assigned to a single split before chunking and question generation. The fixed evaluation pool is a stratified sample of the test split, scored identically by every model and every judge reported in the paper.}
\label{tab:dataset-stats}
\end{table}
\section{Training and Compute Details}
\label{app:training}
This appendix documents the full training configuration used for all mid-stream GRPO runs. Unless a model family is named explicitly, settings are shared across families.
\paragraph{Framework and hardware.}
We align models with GRPO as implemented in TRL, launched with \texttt{accelerate} across $8$ GPUs on a single node ($8\times$NVIDIA B200 180GB). Training runs in \texttt{bf16} with gradient checkpointing enabled.
\paragraph{Optimization.}
We use the \texttt{dr\_grpo} loss with a single GRPO iteration per batch (\texttt{num\_iterations}=1) and $G{=}8$ sampled generations per prompt. The per-device train batch size is $1$ with gradient accumulation of $4$, and we train for one epoch over the mid-stream training set. The learning rate is $1\!\times\!10^{-5}$ under a constant-with-warmup schedule ($10$ warmup steps); we found a constant rate avoided the flat-reward plateau seen with cosine decay during exploration. The KL coefficient against the frozen reference is $\beta{=}0.01$. The maximum completion length is $8{,}192$ tokens, and the maximum prompt length accommodates long medical cases. Each model was trained once, from a single seed; we did not measure seed-to-seed variance.
\paragraph{Parameter-efficient adaptation.}
Only LoRA adapters are trained (rank $r{=}16$, $\alpha{=}64$, dropout $0.05$); base weights stay frozen, and a frozen copy of the backbone serves as the KL reference. Updated adapter weights are periodically re-synchronized to the vLLM engine used for generation.
\paragraph{Sampling.}
Rollout sampling follows each model's recommended settings. For the Qwen families (Qwen3.6-27B, Qwen3.5-9B) and HuatuoGPT-3-8B we use temperature $0.6$, top-$p$ $0.95$, top-$k$ $20$, with a thinking budget of $6{,}000$ tokens for the Qwen families. Reasoning models that emit native thinking tokens are delimited at their closing reasoning tag before the \texttt{<final\_answer>} span is parsed.
\paragraph{Reward and judge.}
The GRPO reward is the four-criterion rubric score (Correct, Specific, Grounded, Complete; weights $3,1,2,1$; normalized to $[0,1]$), produced by DeepSeek-Chat (DeepSeek V4 Flash) at temperature $0.1$ with up to $600$ output tokens. To remove stochastic judge noise on repeated answers, scores are memoized in a shared filesystem cache keyed by a SHA-256 hash of the judge input, so identical answers within a sampling group receive one frozen score. Optional reward shaping (a format bonus and a reasoning-length penalty) is disabled in all reported runs, so the reward equals the normalized rubric score.
\paragraph{Numerical stability.}
For model families that show large per-step policy drift, we clip the log-ratio feeding the K3 KL estimator to bound rare single-token outliers, and for backbones whose chat template ends turns with a token other than the tokenizer's scalar EOS we correct the termination check to the model's true multi-token EOS set. These patches affect training stability only and leave the objective unchanged.
\paragraph{Checkpointing and evaluation cadence.}
We evaluated on the validation split every $100$ GRPO training steps using the same judge protocol and selected the checkpoint with the highest validation accuracy. The results reported in the main paper use \texttt{ckpt-800} for Qwen3.6-27B and \texttt{ckpt-1400} for HuatuoGPT-3-8B.
\begin{table}[t]
\centering
\adjustbox{max width=\columnwidth}{
\begin{tabular}{ll}
\toprule
\textbf{Setting} & \textbf{Value} \\
\midrule
GRPO loss & \texttt{dr\_grpo} \\
Generations per prompt $G$ & 8 \\
GRPO iterations & 1 \\
Per-device batch size & 1 \\
Gradient accumulation & 4 \\
Epochs & 1 \\
Learning rate & $1\times10^{-5}$ (constant + warmup) \\
Warmup steps & 10 \\
KL coefficient $\beta$ & 0.01 \\
Max completion length & 8{,}192 tokens \\
Thinking budget (Qwen) & 6{,}000 tokens \\
LoRA rank / $\alpha$ / dropout & 16 / 64 / 0.05 \\
Precision & bf16 + grad. checkpointing \\
GPUs & 8 (single node) \\
Rollout backend & vLLM (colocated) \\
Sampling & $T{=}0.6$, top-$p$ $0.95$, top-$k$ $20$ \\
Judge model / temperature & DeepSeek-Chat / 0.1 \\
Log / save / eval steps & 5 / 25 / 100 \\
\bottomrule
\end{tabular}}
\caption{Mid-stream GRPO training configuration.}
\label{tab:hparams}
\end{table}
\paragraph{Supervised fine-tuning baseline.}
The \textsc{-sft} checkpoints are trained on the same $18{,}581$ mid-stream training instances, with the generated CoT and answer $(r_i, a_i)$ as the target sequence under a standard next-token objective. They use the same LoRA configuration, precision, and hardware as the GRPO runs.
\section{Prompt Templates}
\label{app:prompts}
This appendix collects the prompt templates used to build and to score MedUPSQA. All are reproduced verbatim; braces mark the fields substituted per instance. The chunking prompt is released with the dataset.
\begin{figure*}[t]
\centering
\begin{tcolorbox}[colback=blue!5!white,colframe=clinicalblue,title={\small Policy system prompt (next-step task)}]
{\scriptsize\ttfamily
You are a clinical reasoning assistant. You will be shown a portion of a patient case and a question whose answer is determined by what comes next in that case --- typically the most appropriate next test, treatment, diagnosis, finding, or clinical decision given the information available so far.\\[2pt]
Follow these rules when answering:\\
- Reason only from the information stated in the case. Do not invent findings, labs, history, or details that are not provided.\\
- Be as clinically specific as the case supports. Name the specific test, drug, anatomical site, or action rather than a vague category --- for example, "CT chest with IV contrast" rather than "imaging".\\
- If the correct answer has multiple components, include all of them.\\
- Think through the case briefly inside <think>...</think>, then output your final answer wrapped in <final\_answer>...</final\_answer> tags.\\[3pt]
\textrm{\textbf{User:}}\\
Case presentation:\\
\{context\_chunk\}\\[2pt]
Question:\\
\{question\}}
\end{tcolorbox}
\caption{\textbf{Policy prompt for the mid-stream next-step task.} The prediction is parsed from the \texttt{<final\_answer>} span. The same template is used for every policy model, aligned and unaligned.}
\label{fig:prompt-policy}
\end{figure*}
\begin{figure*}[t]
\centering
\begin{tcolorbox}[colback=blue!5!white,colframe=clinicalblue,title={\small Evaluation judge (mid-stream next-step task)}]
{\scriptsize\ttfamily
You are a clinical expert evaluating whether a model's answer to a case question matches the reference answer. Consider semantic equivalence and clinical correctness --- phrasing and synonyms do not matter, but the clinical content (action, test, drug, finding, decision, or diagnosis) must be the same.\\[2pt]
Question: \{question\}\\
Model's answer: \{pred\}\\
Reference answer: \{ref\}\\[2pt]
After analysis, conclude with JSON containing:\\
"judgment": 1 if the model's answer is equivalent to the reference, 0 otherwise}
\end{tcolorbox}
\caption{\textbf{Binary equivalence judge for the mid-stream task.} Applied at evaluation time only, identically to every model and every judge family. It is stricter than the four-criterion training rubric.}
\label{fig:prompt-judge}
\end{figure*}
\begin{figure*}[t]
\centering
\begin{tcolorbox}[colback=blue!5!white,colframe=clinicalblue,title={\small Dataset quality validation (construction-time equivalence judge)}]
{\scriptsize\ttfamily
Act as a clinical expert evaluating answer correctness. Decide wether the proposed answer is correct given the reference information containing the ground truth answer.\\
When comparing the two answers, consider:\\[2pt]
1.Accuracy: Match with gold-standard answer. Key errors? (e.g., wrong test referral)\\
2.Completeness: All critical components included?\\
3.Context: Appropriate for this patient's specifics?\\
4.Precision: Specific enough for clinical use? (e.g., "CT abdomen with contrast" vs "imaging")\\[2pt]
For this clinical question, in context:\\
Question: \{question\}\\
Context: \{case\_context\}\\[2pt]
Between this answer and reference information containing the ground truth answer:\\
Proposed Answer: \{final\_answer\}\\
Reference Information: \{reference\_answer\}\\[2pt]
After careful analysis, conclude with JSON containing:\\
"judgment": 1 if correct, 0 if incorrect}
\end{tcolorbox}
\caption{\textbf{Construction-time quality validation prompt.} Applied once per candidate decision point during dataset construction. An instance is retained in MedUPSQA only when the judge returns \texttt{judgment}$=1$; this filter reduces $46{,}447$ candidates to the $21{,}874$ retained instances. Unlike the evaluation judge in Figure~\ref{fig:prompt-judge}, this prompt also sees the accumulated case context, so it can reject answers that are defensible in general but wrong for this patient. The wording was reached iteratively, with $30$ judged instances inspected manually after each revision. Reproduced verbatim, including the original typographical errors.}
\label{fig:prompt-validation}
\end{figure*}
\section{Qualitative Example}
\label{app:examples}
Table~\ref{tab:qualitative} follows one case through three successive decision points, showing at each the question asked, the reasoning trace, the prediction, and the continuation the report recorded.
\begin{table*}[t]
\centering
\setlength{\tabcolsep}{4pt}
\renewcommand{\arraystretch}{1.12}
\scriptsize
\begin{tabular}{@{}S{0.075\linewidth} L{0.283\linewidth} L{0.283\linewidth} L{0.283\linewidth}@{}}
\toprule
\rowcolor{hdrBand}
 & \textbf{Cut 1 \textemdash{} Examination}
 & \textbf{Cut 2 \textemdash{} Identification}
 & \textbf{Cut 3 \textemdash{} Escalation} \\
\midrule
\rlab{Question}
 & What will be the expected findings on corneal examination suggestive of fungal infection?
 & What will be the next diagnostic step to identify the mycelium?
 & What will be the next diagnostic step after the infection spreads to the adjacent sclera and progresses to endophthalmitis? \\
\rowsep
\rlab{Trace}
 & ``Fungal keratitis classically presents with satellite lesions, \hC{feathery margins}, \hB{hypopyon}. But based strictly on the provided text: \hA{subepithelial and superficial stromal} opacities with dot-like and tentacle-like infiltrates \ldots{} the instruction says: reason only from the information stated in the case.''
 & ``\hC{MALDI-TOF-MS} was attempted but failed due to lack of reference spectrum in the database \ldots{} the standard next step for identifying an unknown fungus is molecular sequencing, specifically \hA{ribosomal RNA (rRNA) gene sequencing} (e.g.\ \hB{ITS} region sequencing).''
 & ``When corneal opacity prevents fundus view, \hA{B-scan ultrasonography} is used \ldots{} Sometimes \hC{enucleation} is done, but that's therapeutic/surgical, not diagnostic. The question specifically asks for a \emph{diagnostic} step.'' \\
\rowsep
\rlab{Prediction}
 & \hA{Subepithelial and superficial stromal} opacities with dot-like and tentacle-like infiltrates, often with a central corneal ulcer and \hB{hypopyon} in advanced cases.
 & Molecular identification via \hA{ribosomal RNA (rRNA) gene sequencing}, targeting the \hB{internal transcribed spacer (ITS)} region with panfungal primers.
 & \hA{B-scan ultrasonography (ocular ultrasound)}, essential when direct fundus visualization is obscured; it guides further management (vitreous sampling, vitrectomy, or \hC{enucleation}). \\
\rowsep
\rlab{Report}
 & A grayish-white ulcer measuring $4\times6$\,mm; inflammatory infiltrates with \hC{feathered margins} and \hB{hypopyon} 2\,mm deep; dense central stromal opacity surrounded by a reticular pattern of \hA{subepithelial and superficial stromal} infiltration.
 & The mycelium was sent for \hB{internal transcribed spacer (ITS)} \hA{rRNA gene sequencing} analysis and unambiguously identified; the reference spectrum was then added to the in-house \hC{MALDI-TOF-MS} database.
 & The infection still spread to the adjacent sclera and progressed to endophthalmitis. \hC{Enucleation} eventually had to be done. \\
\rowsep
\rlab{Verdict}
 & \vOK
 & \vOK
 & \vNO \\
\bottomrule
\end{tabular}
\caption{\textbf{One \textsc{MedUPSQA} case cut at three successive decision points} (severe filamentous keratitis). At each cut the model sees the report only up to that point and predicts what happens next, graded free text against free text. Shading links the same clinical entity wherever it recurs within a column. Cut~3 is a graded failure that is not a clinical one: its answer, B-scan ultrasonography, is standard practice when corneal opacity blocks the fundus view, but the report never mentions it.}
\label{tab:qualitative}
\end{table*}
\end{document}